\documentclass[conference]{IEEEtran}
\IEEEoverridecommandlockouts

\usepackage{cite}
\usepackage{amsmath,amssymb,amsfonts}
\usepackage{algorithmic}
\usepackage{graphicx}
\usepackage{textcomp}
\usepackage{xcolor}

\def\BibTeX{{\rm B\kern-.05em{\sc i\kern-.025em b}\kern-.08em
    T\kern-.1667em\lower.7ex\hbox{E}\kern-.125emX}}
\begin{document}

\title{Plantation Monitoring Using Drone Images: A Dataset and Performance Review}


\author{\IEEEauthorblockN{Yashwanth Karumanchi}
\IEEEauthorblockA{
\textit{University of Utah}\\
Salt Lake City, UT, USA\\
u1518595@utah.edu}
\and
\IEEEauthorblockN{Gudala Laxmi Prasanna}
\IEEEauthorblockA{\textit{WASSAN}\\
Hyderabad, India\\
gprasanna@wassan.org}
\and
\IEEEauthorblockN{Snehasis Mukherjee}
\IEEEauthorblockA{
\textit{Shiv Nadar Institution of Eminence}\\
Delhi NCR, India \\
snehasis.mukherjee@snu.edu.in}
\and
\IEEEauthorblockN{Nagesh Kolagani}
\IEEEauthorblockA{
\textit{Centurion University of Technology and Management}\\
Vizianagaram, AP, India \\
nagesh.kolagani@alumni.iitm.ac.in}
}

\maketitle

\begin{abstract}
Automatic monitoring of tree plantations plays a crucial role in agriculture. Flawless monitoring of tree health helps farmers make informed decisions regarding their management by taking appropriate action. Use of drone images for automatic plantation monitoring can enhance the accuracy of the monitoring process, while still being affordable to small farmers in developing countries such as India. Small, low cost drones equipped with an RGB camera can capture high-resolution images of agricultural fields, allowing for detailed analysis of the well-being of the plantations. Existing methods of automated plantation monitoring are mostly based on satellite images, which are difficult to get for the farmers. We propose an automated system for plantation health monitoring using drone images, which are becoming easier to get for the farmers. We propose a dataset of images of trees with three categories: ``Good health", ``Stunted", and ``Dead". We annotate the dataset using CVAT annotation tool, for use in research purposes. We experiment with different well-known CNN models to observe their performance on the proposed dataset. The initial low accuracy levels show the complexity of the proposed dataset. Further, our study revealed that, depth-wise convolution operation embedded in a deep CNN model, can enhance the performance of the model on drone dataset. Further, we apply state-of-the-art object detection models to identify individual trees to better monitor them automatically. The dataset along with annotations, and the codes, will be open sourced for further research.
\end{abstract}


\section{Introduction}
Continuous monitoring of tree health in plantations spread across large areas of villages is necessary for their efficient management. An autonomous system is required to individually monitor trees in such large areas, for reducing human effort and to deal with the difficulty to reach several parts of the plantations regularly.

For this reason, precision agriculture is a necessity now-a-days, where a scientific system allows the farmers to tailor their practices based on the specific needs of different trees in a plantation. Tree health monitoring allows us to efficiently monitor the health of trees regularly, address the problems in case a particular tree is showing unhealthy symptoms, and even manage to prevent the spread of diseases among trees, as and when necessary. Despite the recent advancements in deep learning based models for scene classification or tree health monitoring, there has been a significant lack of efficient deep learning models specific for plantation health monitoring from RGB images. The RGB images captured by UAVs provide detailed and fine-grained information about the individual trees. Additionally, RGB images are more closure to what the human eyes are used to. This provides a comprehensive view of the trees, helping the farmers to comfortably rely on the monitoring system. Moreover, compared to the other types of sensors, RGB cameras are generally more cost-effective, and easy to get by small farmers in India. This makes UAVs equipped with RGB cameras a practical choice for many farmers and researchers interested in automated plantation monitoring.

We aim to create a dataset that can help with the estimation of tree health in a large plantation. The proposed annotated dataset will motivate the researchers to propose deep learning models for identifying individual trees, and estimating their health automatically from drone images. Considering existing literature reviews and pre-trained models, we experiment with several popular CNN models such as AlexNet \cite{alex}, VGGNet \cite{c7}, XceptionNet \cite{xcep}, GoogleNet \cite{google}, EfficientNet \cite{efficient}, and ResNet \cite{c9} models to compare their performances on the proposed dataset.

The motivation behind this study is to develop an efficient and easy-to-use technology to perform individual tree health monitoring in plantations, that could help the farmers with their efficient management. Moreover, the implications of the study would go beyond simple tree monitoring, as the datasets and the performances of the existing CNN models can motivate the researchers to build suitable neural networks specifically designed for plantation monitoring, with minimal cost and efforts by the farmers.

During the last decade, the recent advancements in deep learning models are prompting the researchers to apply CNN models on field images for precession agriculture \cite{c1,c2,c3,c4,c5,c6}. Most of the image-based techniques available in the literature for precession agriculture applications, are based on remote sensing images \cite{c1,c2} for crop and plantation monitoring. However, despite the simplicity of RGB images captured from a UAV, applying deep learning models on images captured from UAVs, remains an unexplored area of research \cite{c3}, although the RGB based techniques provide a low-cost and easy to obtain solution for the farmers. We propose to apply deep learning models such as AlexNet \cite{alex}, VGGNet \cite{c7}, XceptionNet \cite{xcep}, GoogleNet \cite{google}, EfficientNet \cite{efficient}, and ResNet \cite{c9} to explore RGB aerial images captured by our own UAV camera. Further, we create an annotated dataset for tree monitoring, for further studies by the scientific research community.

Panday et al. \cite{c10} attempted to classify RGB drone images for classifying different plants. They designed a CNN model named as the Conjugated Dense Convolutional Neural Network (CD-CNN), harnessing the advantages of a dense architecture. However, CD-CNN is found prone to overfitting on the plant images, due to the smaller number of annotated training samples available for the experiments \cite{cvip}. Increasing the number of training images will increase the cost and manual efforts for annotation, thus reducing the applicability to the farmers.

However, there is no study found in the literature, specifically for the RGB images for tree monitoring. Our contributions are summarized as follows:

\begin{itemize}
    \item We introduce a new dataset of RGB images captured from a camera placed under our own UAV. We use CVAT annotation tool \cite{cvat} to annotate the dataset to release to the research community.
    \item We experiment with several popular CNN models to analyze the nature of the newly introduced dataset.
    \item We further annotate the same dataset for individual tree identification, and apply state-of-the-art models to observe the effect on the new dataset.
    \item Our study reveals that, depth-wise convolution operation can enhance the performance of a model over the proposed dataset.
\end{itemize}

Next, we conduct a survey of related works in the literature. 

\section{Related Works}
During the last decade, the research in precession agriculture has been evolving based on deep learning models \cite{c1,c2}. Deep learning facilitates the automated monitoring of trees in a large plantation, reducing human efforts and time. Several efforts are found in the literature for crop and plantation monitoring using satellite images, e.g., hyperspectral images \cite{afe_23}. Fewer efforts can be found applying deep learning models on UAV-based RGB images \cite{c3}. Yeh et al. applied the two-stage object detection model mask R-CNN to identify the crop area in the image \cite{afe_23_2}. They further applied Alexnet model on the detected regions for classification of crops. Recent studies \cite{c10,cvip} observe the growing importance of RGB images being used in automatic monitoring of plants due to the low cost, flexibility, high mobility, ease of use, and safe operation of unmanned UAVs. Efforts are made applying popular CNN models such as VGGNet \cite{c7}, ResNet \cite{c9}, and other deep networks such as LSTM \cite{c2}, etc., on RGB images. However, no rigorous experiments have been found yet, to observe the effect of the popular CNN models when applied to images of agricultural fields. The existing methods are trained on traditional image classification datasets and applied on agricultural field images. Hence, such models often are not capable of analyzing the minute features obtained from the tree images. A specially designed study for plantation monitoring on RGB images, is necessary, which is absent in the current literature.

There have been a significant number of efforts in using temporal information from the images for the classification of crops \cite{agri23,rse21}. Li et al. \cite{agri23} experimented with popular deep learning models such as 1D-CNN and 2D-CNN. Further, they emphasized on the temporal features using 3D-CNN, LSTM, and some combinations of 1D-CNN and 2D-CNN (for spatial features) with LSTM (for temporal features). According to their reports, 3D-CNN shows the best performance due to its ability to efficiently combine the spatial and temporal features from the crop images. However, such huge 3D-CNN models are computationally heavy. Moreover, such models are extremely data-hungry, making it a difficult choice for plantation monitoring, as the size of the dataset is less. Turkoglu et al. \cite{rse21} combined the spatial and temporal information by proposing a convRNN model. On the same line of thought, Macedo et al. \cite{c2} proposed a convLSTM model to leverage the combination of the spatial and temporal features from crop images.

Efforts have been made to apply transformer models for plant monitoring, taking advantage of the attention mechanism \cite{trans1,trans2}. Wang et al. \cite{trans1} proposed a light-weight transformer model, to classify the nature of disease in infected crops. He et al. \cite{trans2} applied a swin-transformer model for the classification of crop infection based on leaf images. It is observed that 2D-CNNs are more effective compared to the transformers on crop images \cite{c10}. However, finding a suitable CNN model specific to a given set of tree images is challenging. Verma et al. \cite{verma_cea} made an effort, based on a meta-learning approach, to find a suitable CNN for some given plant images.

Most of the existing deep learning-based methods for precession agriculture use remote sensing data, which is difficult to obtain for the farmers. A few efforts have been made to use some other easily available images for plant monitoring \cite{rse24,cea1}. Barriere et al. \cite{rse24} used multimodal data comprising satellite images along with crop rotation patterns and contextual information about the crops. Luo et al. \cite{cea1} used the FPGA data, for plant disease classification, proposing a CNN model. However, UAV-based RGB images are comparatively easier and cheaper for farmers to obtain.
\begin{figure}[!ht]
\centering
\includegraphics[width=0.48\textwidth]{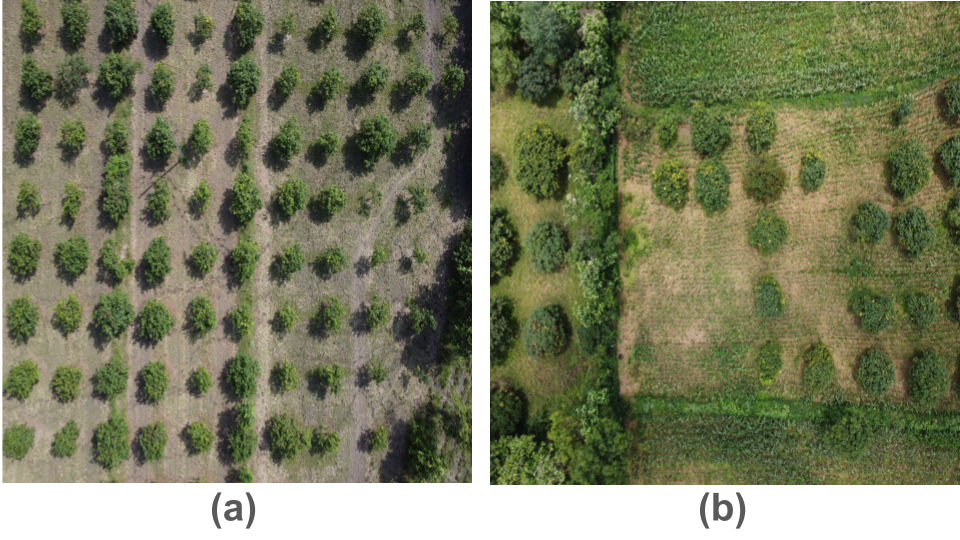}
\caption{Sample drone images in the proposed dataset. These images are annotated using CVAT annotation tool for individual tree identification.}
\label{overall}
\end{figure}

Fewer efforts are found applying CNNs on drone images for plant monitoring, with reduced overfitting \cite{rgb1,trans2}. Joshi et al. \cite{rgb1} used UAV-based RGB images for plant disease classification by a pretrained CNN model, to reduce overfitting. Lin et al. \cite{light} avoided the possibility of overfitting by proposing a lightweight CNN model for crop classification. However, such light-weight models have lesser accuracy on plant images, due to oversimplification of the classifier.

Panday et al. \cite{c10} introduced a dense CNN architecture for plant classification. They introduced a novel activation function named SL-ReLU to address gradient explosion. However, the CD-CNN model proposed by Panday et al. \cite{c10} has not rigorously been tested on RGB drone images.

The goal of this study is to provide a platform for research on automatic plantation health monitoring based on RGB drone images. Next, we discuss the proposed approach.

\section{Proposed Approach}
We perform two different operations on the drone images: individual tree detection and monitoring the health of a tree.

\subsection{Individual Plant Detection}
Detecting individual trees in a drone image can be considered as an application of object detection in computer vision. In a study focusing on a 1-acre mango farm in Tirupati village, we initiated an exploration of YOLO's functionality by dividing a single image into multiple parts. This initial image segmentation resulted in four sub-images, each annotated with 3 tree categorizations—dead, stunted, and good. Subsequently, we embarked on training the YOLO v8 \cite{yolo8} model from scratch, achieving an accuracy of 57.7\%.

Following this, we conducted further experimentation using YOLO v8 \cite{yolo8} with pre-trained weights from the YOLO architecture on the same dataset. Remarkably, the accuracy improved to 69\%. Despite employing only a solitary image, our objective was to comprehensively evaluate the operational dynamics of the YOLO framework.

We further experiment with other YOLO versions, such as YOLO v7 \cite{yolo7} and YOLO v9 \cite{yolo9}, with a pre-trained network, fine-tuned on the proposed dataset.

\subsection{Plant Health Monitoring}
After individual identification of trees, we perform monitoring of the health of the individual plants. For plant health monitoring, we consider 3 classes of plants: ``dead (brown)", ``stunted (light green)", and ``good (dark green)". We apply different state-of-the-art CNN models for classification, such as AlexNet \cite{alex}, VGGNet \cite{c7}, XceptionNet \cite{xcep}, GoogleNet \cite{google}, EfficientNet \cite{efficient}, and ResNet \cite{c9}, and compare the results to analyze their performances.

\section{Dataset and Experiment}
We first provide an overview of the proposed dataset, followed by the experimental set up.

\subsection{Proposed Dataset}
We apply Image Classification Techniques to drone-based data. The dataset annotations were created for individual plant detection boundaries, as well as the class of the plant (dead, stunted, or good). A total of 9,534 annotations (i.e., individual trees) were labeled and cropped using CVAT annotation tool, from 255 drone images. The individual plants are then divided into three categories: dead (1,306 samples), stunted (2,944 samples), and good (5,284 samples). To ensure unbiased training, 1,206 images were randomly selected from each category.
\begin{figure*}[!ht]
\centering
\includegraphics[width=\textwidth]{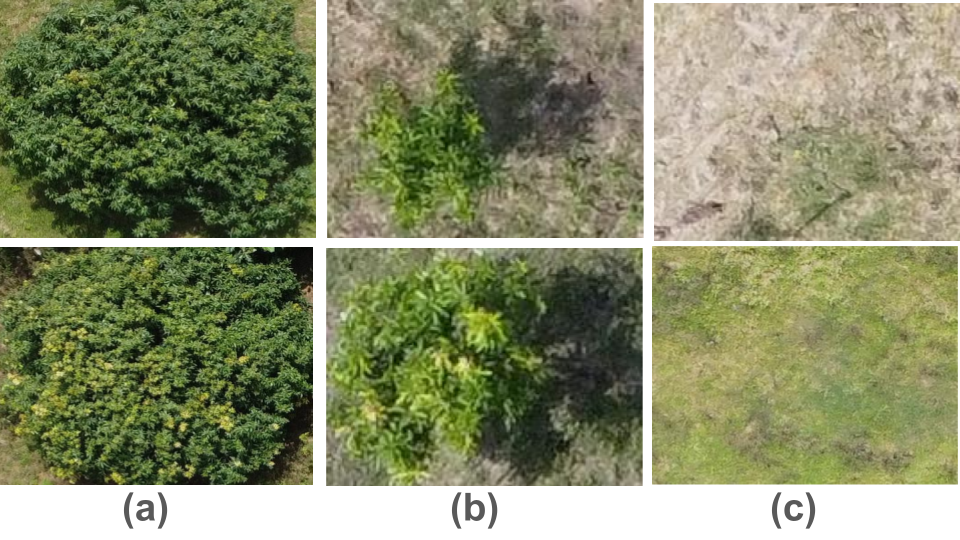}
\caption{Sample annotations performed on the proposed dataset, using the CVAT annotation tool \cite{cvat}. Two samples from each class are shown here: Column (a) Good, Column (b) Stunted, and Column (c) Dead.}
\label{annotation}
\end{figure*}

We fit an RGB camera with a UAV, to capture the field images. The images are captured from a height of 50 meters over the ground. We experimented with different heights and found the 50 meter height as an optimal height for the CNN models to work well for individual plant identification.

Out of a total of 255 images captured by the UAV camera, we randomly select 204 images for training, and the rest 51 images are kept for testing. Figure \ref{overall} shows some sample images captured by the UAV. Size of each image is 5.37Mb, and the resolution of each image is, $4000 \times 3000$. For now, we restrict our study only to Mango trees. Later we will extend our study to generalize our framework for any tree. These 255 images are later annotated to produce the number of annotations shown in the previous paragraph. We apply CVAT annotation tool to annotate the images \cite{cvat}. Figure \ref{annotation} shows some sample annotations on our dataset, with two examples from each class: dead, stunted, and good.

Next, we illustrate the experimental setup for this study.

\subsection{Experimental Setup}
We train multiple pre-trained Image Classification models on this balanced dataset, augmenting it with additional layers for horizontal flip and random rotation. Fully connected layers are appended for final classification, utilizing ReLU activation and SoftMax function. The models are trained with a learning rate of 0.0001, sparse categorical cross-entropy loss, and 100 epochs. The models evaluated include AlexNet, VGG19, GoogleNet, EfficientNet, VGG16, ResNet50, and XceptionNet. In order to cope with the limited training data, we tried light-weight models (AlexNet and VGG16), a deeper model for better accuracy (VGG19, GoogleNet, and EfficientNet), CNN model with Residual connections to avoid overfitting (ResNet50), and a CNN model with depth-wise Convolutions, to reduce the number of parameters (XceptionNet).

Next, we discuss the results of our experiments on the proposed dataset.

\section{Results and Discussions}
We perform the experiments on the proposed dataset, with some popular CNN models. We further analyze the reason behind the performances of the CNN models. The results of applying different CNN models on the proposed dataset for the three-class classification is shown in Table \ref{res}.

\begin{figure*}
\centering
\includegraphics[width=\textwidth]{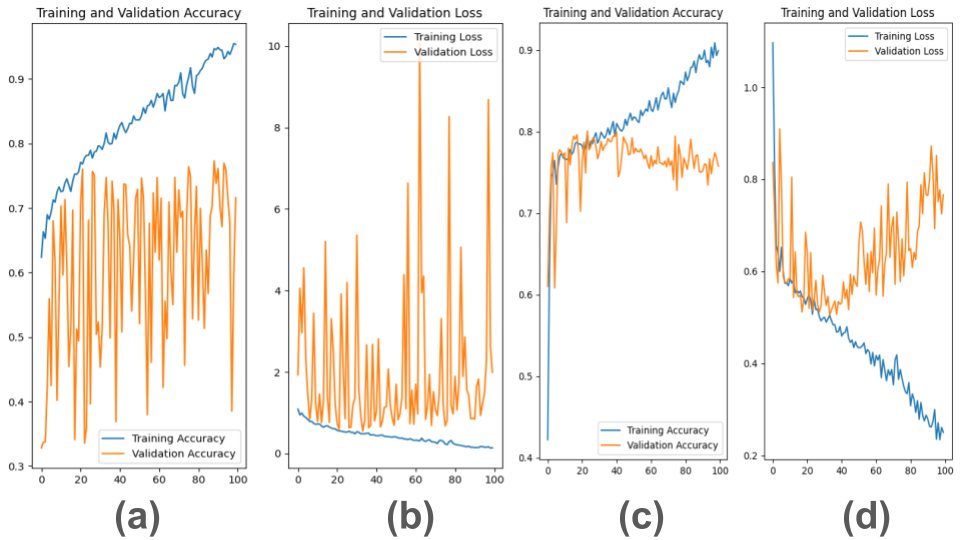}
\caption{Train versus test accuracy and loss curves for AlexNet and VGG19 models: (a) AlexNet accuracy curve, (b) AlexNet loss curve, (c) VGG19 accuracy curve, (d) VGG19 loss curve. The huge differences between the train and test accuracy and loss indicate overfitting by both the models.}
\label{alex_vgg19}
\end{figure*}

We can observe in Table \ref{res} that, shallower networks such as AlexNet, VGG16, and VGG19 could not perform very well due to two possible reasons. First, for a challenging dataset like the proposed drone image dataset, a more deeper networks are necessary for classification. Second, although they are shallow networks, still they are suffering from overfitting problem, due to a very small size of the proposed dataset. Figure \ref{alex_vgg19} shows the train versus test loss curves and accuracy curves for the AlexNet and VGG19 models respectably. We can observe that, both the models are overfitting for the proposed dataset.
\begin{table}
    \begin{center}
        \caption{Performance (in terms of percentage of accuracy) of the different CNN models when applied to the proposed dataset. Further, the execution times are also shown for each model.}\label{res}
        \begin{tabular}{|c|c|c|c|} \hline
            CNN Model & Accuracy (\%) & Loss & Execution Time \\ \hline
            AlexNet \cite{alex} & 71.6 & 1.992 & 13 minutes \\ \hline
            VGG19 \cite{c7} & 75.8 & 0.766 & 1 hour 29 minutes \\ \hline
            GoogleNet \cite{google} & 77.1 & 1.367 & 55 minutes \\ \hline
            EfficientNet \cite{efficient} & 78.3 & 1.540 & 56 minutes \\ \hline
            VGG16 \cite{c7} & 78.9 & 0.962 & 1 hour 14 minutes \\ \hline
            ResNet50 \cite{c9} & 95.6 & 0.153 & 1 hour 10 minutes \\ \hline
            ExceptionNet \cite{xcep} & 97.1 & 0.103 & 1 hour 41 minutes \\ \hline
        \end{tabular}
    \end{center}
\end{table}

\begin{figure*}
\centering
\includegraphics[width=\textwidth]{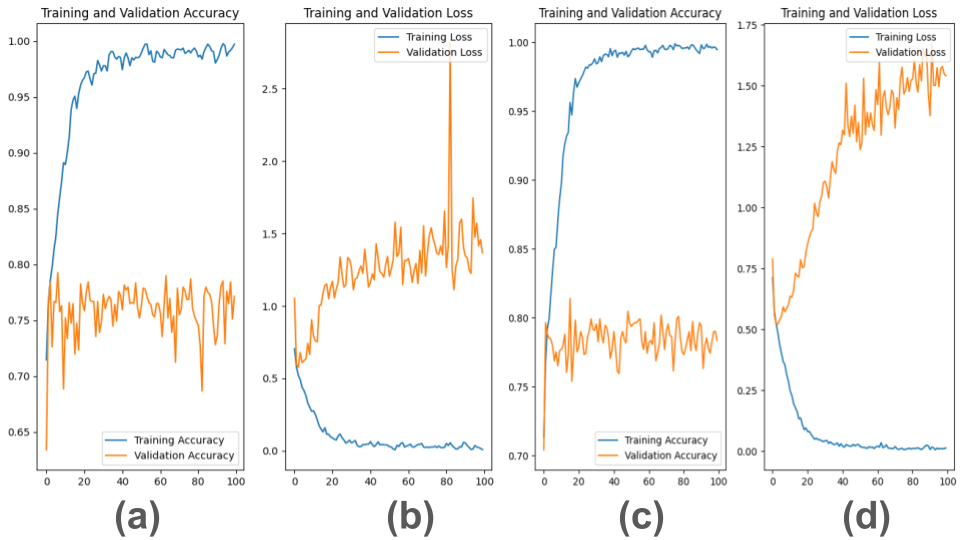}
\caption{Train versus test accuracy and loss curves for GoogleNet and EfficientNet models: (a) GoogleNet accuracy curve, (b) GoogleNet loss curve, (c) EfficientNet accuracy curve, (d) EfficientNet loss curve. The huge differences between the train and test accuracy and loss indicate overfitting by both the models.}
\label{google_efficient}
\end{figure*}

The train versus test accuracy and loss curves are shown for the models GoogleNet and EfficientNet in Figure \ref{google_efficient}. Being very deep CNN models, they are capable of extracting minute features from the images. However, the complexity of the models make them prone to overfitting, which are depicted in Figure \ref{google_efficient}.

\begin{figure*}
\centering
\includegraphics[width=\textwidth]{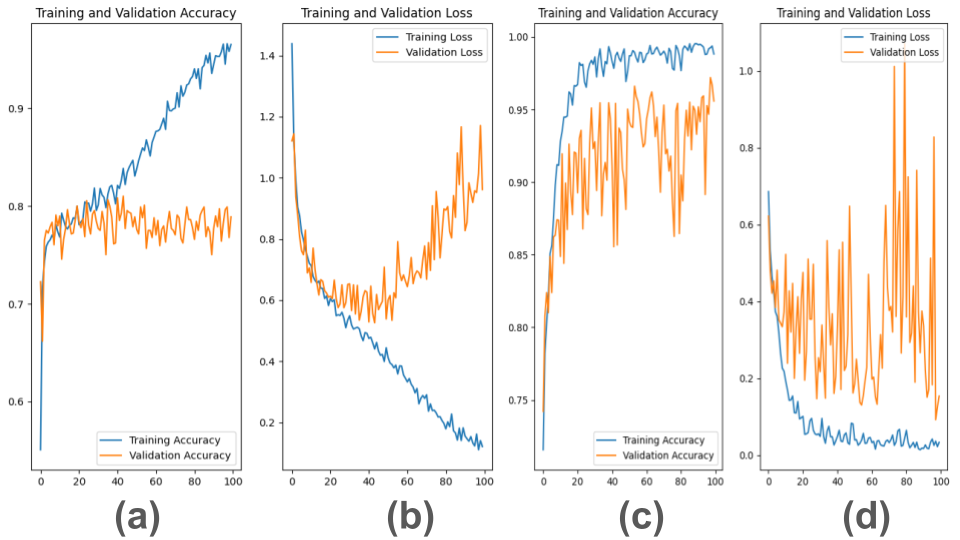}
\caption{Train versus test accuracy and loss curves for VGG16 and ResNet50 models: (a) VGG16 accuracy curve, (b) VGG16 loss curve, (c) ResNet50 accuracy curve, (d) ResNet50 loss curve. The differences between the train and test accuracy and loss is less for VGG16, and even lesser for ResNet50 model.}
\label{vgg16_resnet50}
\vspace{-0.2in}
\end{figure*}

Figure \ref{vgg16_resnet50} shows the train versus test loss curves and accuracy curves for the VGG16 and ResNet50 models respectably. We can observe that, the amount of overfitting is slightly less in case of VGG16, may be due to the shallower structure. However, Table \ref{res} shows a less accuracy for VGG16, possibly due to the simple structure, incapable of capturing the minute texture of the images. We further can observe in Table \ref{res} that, ResNet50, although is a deep network, performed much better compared to the other CNN models, possibly due to the Residual connections, which helped in reducing the overfitting. Figure \ref{vgg16_resnet50} also supports the observation of lesser overfitting by the ResNet50 model. Further, we experimented with ResNet 18 model as well. However, ResNet18 model provided very less accuracy, and hence, we have not included the results in this study. Although ResNet18 is a shallower model with residual connections, it was not found suitable for the drone images, possibly due to the complexity of the drone images.

\begin{figure}
\centering
\includegraphics[width=0.48\textwidth]{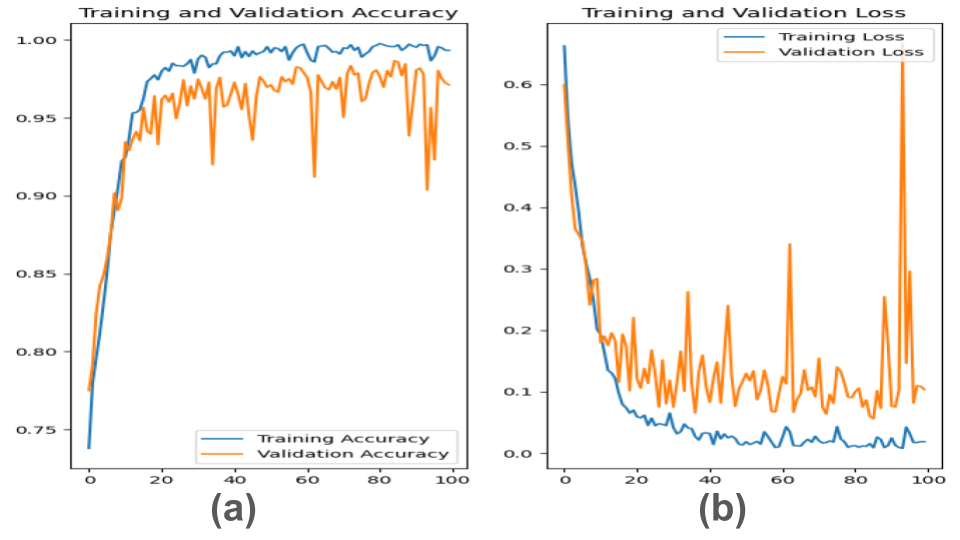}
\caption{Train versus test accuracy and loss curves for XceptionNet model: (a) XceptionNet accuracy curve, (b) XceptionNet loss curve. The difference between train and test accuracies and losses are lesser compared to the other CNN models, indicating lesser overfitting.}
\label{xcepnet}
\end{figure}

Figure \ref{xcepnet} shows the train versus test loss curves and accuracy curves for the XceptionNet model. Clearly, XceptionNet shows a lesser difference between train and test accuracies and losses, indicating lesser overfitting. Moreover, XceptionNet shown the highest accuracy among all the CNN models, according to Table \ref{res}. This shows that, the concept of depth-wise convolution operation helped in enhanced performances even on a small dataset. Based on this observation, in future, research can be conducted with various image classification models, after incorporating the depth-wise convolution into the structure, to further explore on even better CNN models specific to the problem of crop and plantation monitoring.

\section{Conclusions}
The goal of this study was to provide a benchmark for research in plantation health monitoring from RGB images captured by UAV cameras. We propose a challenging annotated dataset which is publicly available for research purpose. We have experimented with seven popular CNN models with sufficient varieties in structure, in order to establish a benchmark for future research on such images. According to our study, depth-wise convolution operation plays an important role in tree health monitoring, may be due to the lesser number of parameters. In future, more suitable CNN models can be built to handle such images, based on deeper architecture and depth-wise convolution operations incorporated in it. Further, some other recent parameter reduction techniques can be incorporated in future, for analyzing plantation images captured by drones. Moreover, the proposed dataset can be used for individual tree monitoring in drone images.

\bibliographystyle{IEEEtran}
\bibliography{cas-refs_new}

\end{document}